\documentclass{article} 
\usepackage{iclr2022_conference,times}

\usepackage{amsmath,amsfonts,bm}

\def\eqref#1{equation~\ref{#1}}

\def\1{\bm{1}}

\DeclareMathAlphabet{\mathsfit}{\encodingdefault}{\sfdefault}{m}{sl}
\SetMathAlphabet{\mathsfit}{bold}{\encodingdefault}{\sfdefault}{bx}{n}

\usepackage{hyperref}
\usepackage{url}
\usepackage{graphicx}
\graphicspath{ {./images/} }
\usepackage[normalem]{ulem}
\useunder{\uline}{\ul}{}
\title{KGRefiner: Knowledge Graph Refinement for Improving Accuracy of Translational Link Prediction Methods}

\author{Mohammad Javad Saeedizade, Najmeh Torabian \& Behrouz Minaei-Bidgoli  \\
Iran University of Science and Technology \\
School of Computer Engineering\\
Data Mining Lab\\
\texttt{\{m\char`_saeedizade,b\char`_minaei\}@iust.ac.ir} \\
\texttt{Najmeh.torabian@gmail.com} \\
}

\iclrfinalcopy 
\begin{document}

\maketitle

\begin{abstract}
The Link Prediction is the task of predicting missing relations between entities of the knowledge graph. Recent work in link prediction has attempted to provide a model for increasing link prediction accuracy by using more layers in neural network architecture. In this paper, we propose a novel method of refining the knowledge graph so that link prediction operation can be performed more accurately using relatively fast translational models. Translational link prediction models, such as TransE, TransH, TransD, have less complexity than deep learning approaches. Our method uses the hierarchy of relationships and entities in the knowledge graph to add the entity information as auxiliary nodes to the graph and connect them to the nodes which contain this information in their hierarchy. Our experiments show that our method can significantly increase the performance of translational link prediction methods in H@10, MR, MRR.
\end{abstract}

\section{Introduction}
Knowledge graphs represent a set of interconnected descriptions of entities, including objects, events, or concepts. These graphs are structures by which knowledge is stored in triples. These triples include the three parts head, relation, and tail. The relation determines the type of relationship between head and tail. These graphs are becoming a popular approach to display and model different information in the world. Additionally, knowledge graphs have several applications, for example, question answering systems \citep{qa_kg_1,qa_kg_2}, recommendation systems \citep{kb-recommender}, search engines \citep{xiong2017explicit}, relationship extraction \citep{distant_supervision2009}, etc. 

\indent Despite many efforts to build knowledge graphs, they are not complete yet. For example, in the Freebase \citep{freebase}, over 70\% of people do not have their place of birth in the graph. This incompleteness of knowledge graphs has motivated researchers to add information to the graph and complete it.

\indent One of the developing fields in completing the knowledge graph is knowledge graph embedding (KGE). The task of KGE is to embed entities and relationships in a small continuous vector space. One application of these embedding is to predict missing links in the knowledge graph.

Translational link prediction models use the sum of the head and relation vectors to predict the tail. These models started with TransE \citep{bordes2013translating}, and after that, TransH \citep{transh}, TransR \citep{lin2015learning}, TransD \citep{ji2015knowledge}, RotatE \citep{rotate}, etc., tried to improve it in the following years. The advantages of translational methods over deep learning techniques are that they are robust, and their score function is considerably faster. Therefore, in this work, we tried to improve these translational methods.

\indent There is a lot of information in knowledge graphs. The hierarchy of entities and relationships is part of it. Paris, for example, its hierarchy is ``entity$\,\to\,$physical\_entity$\,\to\,$object$\,\to\,$location$\,\to\,$region$\,\to\,$area$\,\to\,$center$\,\to\,$seat$\,\to\,$capital$\,\to\,$national\_capital''. This hierarchy is not given enough attention in link prediction methods, and we intend to use this information in this paper.

\indent SACN \citep{sacn_paper} added some nodes and relationships to the graph to use the graph structure information but did not justify adding these nodes and edges, so it is not generalizable for other graphs. In addition, SACN added this information only to FB15K237 and did not provide a method for WN18RR. In this paper, we added a much smaller number of relationships and fewer nodes to the graph training section by interpreting them. HRS \citep{zhang2018knowledge} used relation clusters and sub-relations to use this information. Nevertheless, like SACN, this can not be generalized well.

The \citep{moon2017learning} considered that if two entities are embedded closely in the embedding space, they are similar and assigned entities' classes based on closeness. Still, we assumed that if two entities use the same relation in the graph or have common elements in their hierarchies, they are related. 

When link prediction models learned the relation between Paris and France, previous link prediction methods did not notice that Paris is a city and France is a country. To use this information, we added auxiliary nodes to the graph that included the classes of entities and connected them to related entities. For example, we added an extra node for countries to the knowledge graph and connected it to all the knowledge graph countries. Our contributions are as follows:
\begin{enumerate}
	\item[$\bullet$]  We presented a method for refining the knowledge graph, which is independent of the structure of the link prediction model and adds triples to the knowledge graph. These triples increase the accuracy of link prediction with the same time and space complexity of translational models.
	\item[$\bullet$]  We evaluated our proposed method on two FB15K237 and WN18RR datasets with successful translational models. The results showed that accuracy in link prediction was significantly increased on H@10, MRR, and MR.
  \end{enumerate}
\section{Related Work}
Knowledge graph embedding is an active and developing field to embed the entities and relations of the knowledge graph. These embeddings are used in link prediction, question answering systems, relation extraction, etc. Knowledge graph embedding starts with TransE \citep{bordes2013translating}, which is the first translational link prediction method. It interprets relation as a transition from head entity to tail in the graph. Some drawbacks of the TransE model are its inability to model N-1, 1-N, and N-N relationships. In the following years, some other translational approaches, such as TransH \citep{transh}, TransD \citep{ji2015knowledge}, and TransR \citep{lin2015learning}, were inspired by the initial idea of TransE \citep{bordes2013translating} and tried to improve it. These translational models have much more speed against deep learning models such as ConvE \citep{conve}, ConvKB \citep{convkb}, SACN \citep{sacn_paper}, and HAKE \citep{zhang2020learning}, but their accuracy is slightly lower than these models. Therefore, we proposed a method to increase the accuracy of these translational models.
\\
\indent
Knowledge graph refinement is a field of correcting or improving the knowledge graph. BioKG \citep{zhao2020biomedical}, which worked on medical graphs, has tried to provide a method for removing the wrong information in these graphs. Other works in the refinement of the knowledge graphs try to add information. SACN \citep{sacn_paper} has also added attributes to the knowledge graph, like our work. SACN proposed FB15k237\_Attr; this method for constructing this dataset has three major issues. First, it only worked for FB15k237, but our proposed method can be applied on WN18RR as well. Second, it has brought the number of FB15k237 relations from 237 to 484; therefore, it has more time complexity than ours. However, we only proposed two new relations for FB15k237 and only one relation for WN18RR. Third, these new relations and entities are not interpretable in SACN; It does not provide a reason for adding these attributes. So it can not be generalized on other graphs. \\
\indent
HRS \citep{zhang2018knowledge} tried to use sub-relation and relation-cluster to make better predictions. It used the hierarchy of relations as a sub-relationship, and it created a relation cluster to use these as two additional parts of the transition in the translational models. Because links in Wordnet do not have information about entities, HRS sub-relation and relation-cluster on Wordnet are meaningless.
\section{Background}
\label{gen_inst}
\noindent Suppose E as the collection of all entities of knowledge graph and R set of all its relationships. The ($e_s$, $r$, $e_o$) is called a triple. The $e_s \sim$ E is the head, and $e_o \sim E$ is the tail of a triple. Finally, $r \sim E$ represents the relation between $e_s$ and $e_o$.
\subsection{Link Prediction}
Link prediction is the task of predicting the missing link of a knowledge graph by inferring from existing facts on it. The score function of link prediction methods is $\psi(e_o, r, e_s)$, which evaluates triple's accuracy. Our goal in teaching a model that has the highest estimation for the missing triplets of the graph and the lowest prediction for false triples.
\subsection{Translational Link Prediction Models}
Translational link prediction methods consider the relation as a transition from head to tail. For example (Paris, Capital of, France), the relation ``Capital of'' is a transition from Paris to France. TransE \citep{bordes2013translating} is the first translational link prediction model. In TransE, embeddings for correct triples are learned as $e_s + r \sim e_o$. It means that the sum of the head's embedding and relation's embedding must be close to the tail; primarily, the distance measure is the L2 norm.
Here are some translational link predictions:\\

\noindent \textbf{TransE:}
For factual triple ($e_s$,$r$,$e_o$), adding embeddings of head and relation should be closed to the tail embedding, and on the other hand, for corrupted ones ($e_s$,$r$,$e_o\prime$),  $e_s +  r$ should have a distance with $e_o\prime$. The score function of TransE is as follow: 
\begin{align*}
    \psi(e_o, r, e_s) &= -|| h + r - t ||^2_2
\end{align*}
\noindent \textbf{TransH} \citep{transh}\textbf{:}
To improve modelling of N-1, 1-N and N-N, TransH defined a hyperplane for each relations, and translation property should be established on that hyperplane.
\begin{align*}
    h_ \bot = w^ \bot_r hw_r \,\, &,\,\, t_ \bot = w^ \bot_r tw_r \\
    \psi(e_o, r, e_s) &= -|| h_ \bot + r - t_ \bot ||^2_2
\end{align*}

\noindent \textbf{TransD} \citep{ji2015knowledge} \textbf{:}
It creates a dynamic matrix for all entity-relation pairs and maps the head and tail into M1 and M2, respectively. The transition from head to tail is as follow:
\begin{align*}
    M^1_r  =  w_r w^ \bot_h + I \,\, &,\,\, M^2_r  =  w_r w^ \bot_t + I \\
    h_ \bot = M^1_r h \,\, &,\,\, t_ \bot = M^2_r t \\
    \psi(e_o, r, e_s) = -|&| h_ \bot + r - t_ \bot ||^2_2
\end{align*}
\noindent \textbf{TransR} \citep{lin2015learning} \textbf{:}
It considers that entities may have multiple aspects, and various relations focus on different aspects of entities. It projects entities into relation space by projection matrix M.
\begin{align*}
    h_ \bot = M_r h \,\, &,\,\, h_ \bot = M_r t  \\
    \psi(e_o, r, e_s) = -|&| h_ \bot + r - t_ \bot ||^2_2
\end{align*}
\noindent \textbf{RotatE} \citep{rotate} \textbf{:}
RotatE deals with relation as a rotation to complex space. This rotation brings the source entity to the target entity in the complex space. The relation applies to the head entity by  Hadamard product. Then it uses the L1 norm to measure the distance from the tail entity in the score function.
\begin{align*}
    \psi(e_o, r, e_s) &= -|| h_ \circ r - t_ \bot ||^2
\end{align*}

\subsection{Knowledge Graph Refinement}
The knowledge graph refinement follows two main objectives: (A) adding information to the knowledge graph, which is a subcategory of the knowledge graph completion. (B) Detecting incorrect information and remove those triplets from the knowledge graph to increase the correctness of the knowledge graph.


\section{KGRefiner}
\label{headings}
\begin{figure*}[ht!]
	\centering
			\includegraphics[width=1\textwidth]{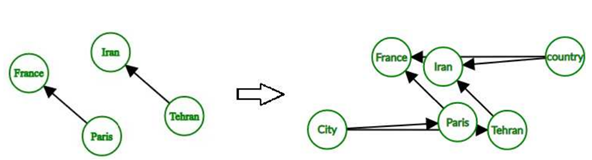}
		\label{fig:verticalcell}
		\caption{Simple illustration of changes in embedding space. The right side graph shows the effect of adding auxiliary nodes to the graph, which translational models bring all countries together and cities together in vector space.}
\end{figure*}
In this work, we propose a method to add information to the graph, which refines the knowledge graph and increases link prediction accuracy. In FB15k237,  we do this refinement by using relation hierarchies, and in WN18RR, we use hierarchies of entities. We add this information to the graph as a new node; these nodes are auxiliary nodes. We introduce several new relations to connect these new nodes to graph nodes, and we add these triples to the graph.\\ 
Translational link prediction methods such as TransE \citep{bordes2013translating}, TransH \citep{transh}, TransD \citep{ji2015knowledge}, etc., create transition property in their embeddings. For example, in TransE, embeddings are made as follow: 
\begin{align}
	e_s + r &\approx e_o
 \end{align}
 This means in embedding space; the tail entity should be close to the sum of head and relation. For example, let's consider these triples: 
 \begin{align}
	Paris + capital of &\approx France \\
	Tehran + capital of &\approx Iran
 \end{align}
Link prediction model is not aware of both tails entities are country. If we add new node as ``country'' to the graph and connect it to all graph's countries with a new relation ``RelatedTo'' then these triples are added to graph: 
\begin{align}
France + RelatedTo &\approx country \\
Iran + RelatedTo &\approx country 
\end{align}
Equations 4 and 5, which are similar, bring closer the embeddings of France and Iran, which are semantically identical. Figure 1 gives an illustration of what changes KGrefiner brings for the embedding space. This closeness in evaluating Equation 2 causes the model to search between countries when asked where France’s capital is.
\subsection{Refinement of FB15k237}
In FB15k237, graph relations contain information about entities. For example, the ``entity$\,\to\,$physical\_entity$\,\to\,$object$\,\to\,$location$\,\to\,$region$\,\to\,$area$\,\to\,$center$\,\to\,$seat$\,\to\,$capital$\,\to\,$national\_capital'' is a relationship between countries and cities, and nodes on one side of relationships can be considered similar. Higher levels usually have more general information about objects in the hierarchy, and lower levels have more specific, so we extracted the last three levels of hierarchies from each relation in this graph to use this information. Then, for each sub-relation, we counted the number of repetitions in the graph training section. We removed those components with less than 100 repetitions in the graph to reduce the number of these sub-relations, and the number 100 is arbitrary. Finally, 285 sub-relations remained, which we added to the set of entities in this graph (as new nodes). We call these auxiliary nodes relation-nodes. We defined two new relations, ``RelatedTo'' and ``HasAttribute'', to connect these relation-nodes to the graph. For each triple, if the entity is the triple's head, we linked it with relation-node by ``RelatedTo'', and if it is the tail of the triple, we use ``HasAttribute'' to establish these connections. For example, to refine relation between Paris and France, (Paris,``entity$\,\to\,$physical\char`_entity$\,\to\,$object$\,\to\,$location$\,\to\,$region$\,\to\,$area$\,\to\,$center$\,\to\,$seat$\,\to\,$capital$\,\to\,$national\char`_capital'',France), ``capital'' has repetition over 100, so the following triples were added to the graph:
\begin{align*}
	France + HasAttribute &\approx capital \\
	Paris + RelatedTo &\approx capital 	
\end{align*}
\subsection{Refinement of WN18RR}
To refine this graph, we use the hierarchy of entities. In Freebase, we used relationships, but relationships do not give us information about entities in Wordnet. France, for example, has a hierarchy of ``existence $\,\to\,$ place $\,\to\,$ region $\,\to\,$ region $\,\to\,$ administrative region $\,\to\,$ country $\,\to\,$ France''. This hierarchy gives us good information about France. Except for the last level, we extract the other last three levels of entities.  Among these levels, we hold those with more than an arbitrary number of 50 repetitions among entities to reduce these levels. As a result, 207 levels remained. We add these levels as new nodes to the graph training section and connect them to the entities with these levels in their hierarchy with a new type of connection. In this graph, we define a new relation and name it ``HasAttribute''. For example, France and Iran have a ``country'' in their hierarchical structure. Then, the following triples were added to the training section of the graph:
\begin{align*}
	France + HasAttribute &\approx country \\
	Iran + HasAttribute &\approx country 	
\end{align*}
\begin{table*}[h]
  \begin{center}
    \begin{tabular}{l|c|c|c|c} 
      \textbf{Dataset} & \textbf{FB15k237} & \textbf{FB15k237-Refined}& \textbf{WN18RR} & \textbf{WN18RR-Refined}\\
      \hline
      Entities & 14541 & 14826& 40943& 41150\\
      Relations & 237 & 239& 11& 12\\
      Train Edges & 272115 & 550998& 86835& 230135\\
      Val. Edges & 17535 & 17535& 3034& 3034\\
      Test Edges & 20466 & 20466& 31134& 31134\\
    \end{tabular}
    \caption{Statistics of the experimental datasets. The refined version represents that graph has some auxiliary nodes.}
\bigskip
\begin{tabular}{|c|ccc|}
\hline
Baseline                    & H@10                & MR                 & MRR                 \\ \hline
TransE                      & 45.6                & 347                & \textbf{29.4}                \\
\textbf{TransE + KGRefiner} & \textbf{47}         & \textbf{203}       & 29.1                \\ \hline
TransD                      & \textbf{45.3}       & 256                & \textbf{28.6}       \\
\textbf{TransD + KGRefiner} & 43.7                & \textbf{227}       & 24                  \\ \hline
RotatE                      & \textbf{47.4}       & {\ul \textbf{185}} & \textbf{29.7}       \\
\textbf{RotatE + KGRefiner} & 43.9                & 226                & 27.9                \\ \hline
TransH                      & 36.6                & 311                & 21.1                \\
\textbf{TransH + KGRefiner} & {\ul \textbf{48.9}} & \textbf{221}       & {\ul \textbf{30.2}} \\ \hline
\end{tabular}
 \caption{Link prediction results on FB15K237 and its refined version. Results of TransE is taken from \protect\citep{convkb}, TransH and TransD from \protect\citep{zhang2018knowledge}, but for RotatE we used \protect\citep{han2018openke} to produce scores.}
\begin{tabular}{|c|ccc|}
\hline
Baseline                    & H@10                & MR                 & MRR                 \\ \hline
TransE                      & 50.1                & 3384               & \textbf{22.6}       \\
\textbf{TransE + KGRefiner} & \textbf{53.7}       & \textbf{1125}      & 22.2                \\ \hline
TransH                      & 42.4                & 5875               & 18.6                \\ 
\textbf{TransH + KGRefiner} & \textbf{51.4}       & \textbf{1534}      & \textbf{20.8}       \\ \hline
TransD                      & 42.8                & 5482               & 18.5                \\
\textbf{TransD + KGRefiner} & \textbf{52.3}       & \textbf{1348}      & \textbf{21.4}       \\ \hline
RotatE                      & 54.7                & 4274               & {\ul \textbf{47.3}} \\
\textbf{RotatE + KGRefiner} & {\ul \textbf{57.0}} & {\ul \textbf{683}} & 44.8                \\ \hline
\end{tabular}

  \caption{Link prediction results on WN18RR and its refined version. Results of TransE is taken from \protect\citep{convkb}, TransH and TransD from \protect\citep{zhang2018knowledge}, for RotatE we used \protect\citep{han2018openke} to produce scores. For other results, we used \protect\citep{han2018openke} to produce them.}

\end{center}
\end{table*}

\section{Exprement}
\subsection{Datasets}
We evaluated our work on popular benchmarks: FB15K237 and WN18RR; these datasets are respectively refined from real knowledge graphs: WordNet \citep{wordnet} and Freebase \citep{freebase}. In addition, we built two other datasets with KGRefiner: FB15K237-Refined and WN18RR-Refined, respectively, from FB15K237  and WN18RR. The details of the datasets are shown in Table 1.

\subsection{Baselines}
To demonstrate the effectiveness of our models, we compare results with the original translational models TransE \citep{bordes2013translating}, TransH \citep{transh}, TransD \citep{ji2015knowledge}, and the last translational model, RotatE \citep{rotate}.

\subsection{Experimental Settings}
We used implementation of baselines by OpenKE \citep{han2018openke}. We used an embedding dimension of 200 for all models. Also, we removed self adversarial negative sampling from TransE and RotatE to have a fair comparison. We tried \{200, 500, 1000, 2000\} epochs, and we picked the best epoch according to MRR on the validation set. Other hyperparameters of the models are those mentioned in OpenKE. Hyperparameters for FB15K237 and FB15K237-Refined and also WN18RR and WN18RR-Refined are the same.

\subsection{Experimental Results}
Table 2 and 3 compares the experimental results of our KGRefiner plus translational models and with previously published results. Results in bold font are the best results in the group, and the underlined results denote the best results in the column. KGRefiner with TransH obtains the highest H@10 and MRR on FB15k237, and also KGRefiner with RotatE reached the best MR and H@10 in WN18RR.
\begin{center}
\begin{table}[]
  \centering
  \begin{tabular}{l|c}
  Model                                    & \begin{tabular}[c]{@{}c@{}}Training Time \\ for single epoch\end{tabular} \\ \hline
  TransE \citep{bordes2013translating} $[\oplus]$                    & 2.8 s                                                                                \\ \hline
  TransH \citep{transh}  $[\oplus]$                   & 5.2 s                                                                                \\ \hline
  TransD \citep{ji2015knowledge}  $[\oplus]$          & 5.2 s                                                                                \\ \hline
  RotatE \citep{rotate}    $[\oplus]$                 & 5 s                                                                                  \\ \hline
  ConvE \\ \citep{conve}     $[\ominus]$                  & 279 s                                                                                \\ \hline
  ConvKB \\ \citep{convkb}   $[\ominus]$                  & 40 s   
                                                                             
  \end{tabular}
  \caption{Comparison between translational technique and deep learning methods in training time. [$\oplus$]: These models are implemented by OpenKE \protect\citep{han2018openke} and [$\ominus$] are produced by their original implementations.}

  \end{table}
\end{center}
\subsection{Speed of Models}
The training time of translational models is much less than deep learning approaches such as ConvE, SACN, ConvKB, etc. The complexity in scoring function and neural network layers in their architecture reduces training speed in deep learning methods. Table 4 compares the time that each model needs to be trained for one epoch on FB15k237. We ran models on Nvidia K80. For fair comparison embedding dimension for all models is 200. These models usually need 1000 epochs, so the runtime difference between TransE and RotatE is around 35000s for FB15k237.


\section{Conclusion}
In this paper, we propose KGRefiner, a novel knowledge graph refinement method that alleviates the limitations of translational models by capturing additional information in knowledge graph hierarchies. We used hierarchy components as new nodes, and by connecting these nodes to proper entities in the knowledge graph, we have a more informative graph. Our experimental results show that our KGRefiner outperforms other state-of-the-art translational models on two benchmark datasets WN18RR and FB15k237. Furthermore, it is the first augmentation method that works with both Wordnet and Freebase, while old methods only perform only on one dataset.
\indent In future works, we will expand our work on datasets that can be formulated on the triple structure. For example, recommender system datasets can be formed on graph schema, and KGRefiner can be applied.

\bibliography{iclr2022_conference}
\bibliographystyle{iclr2022_conference}

\end{document}